\def\year{2022}\relax
\documentclass[letterpaper]{article} 
\usepackage{aaai22}  
\usepackage{times}  
\usepackage{helvet}  
\usepackage{courier}  
\usepackage[hyphens]{url}  
\usepackage{hyperref}
\usepackage{graphicx} 
\usepackage{natbib}  
\usepackage{caption} 
\usepackage{times,multirow,booktabs} 
\usepackage{amsmath,latexsym,amsthm,mathtools,xspace} 
\usepackage{comment} 
\usepackage{algorithm,algpseudocode} 
\DeclareCaptionStyle{ruled}{labelfont=normalfont,labelsep=colon,strut=off} 
\usepackage{enumitem}

  \newlist{compactitem}{itemize}{3}
  \setlist[compactitem]{topsep=0pt,partopsep=0pt,itemsep=0pt,parsep=0pt}
  \setlist[compactitem,1]{label=\textbullet} 
  \setlist[compactitem,2]{label=---} 
  \setlist[compactitem,3]{label=*}

  \newlist{compactdesc}{description}{3}
  \setlist[compactdesc]{topsep=0pt,partopsep=0pt,itemsep=0pt,parsep=0pt}

  \newlist{compactenum}{enumerate}{3}
  \setlist[compactenum]{topsep=0pt,partopsep=0pt,itemsep=0pt,parsep=0pt}
  \setlist[compactenum,1]{label=\arabic*}
  \setlist[compactenum,2]{label=\alph*}
  \setlist[compactenum,3]{label=\roman*}

\title{On the Use of Unrealistic Predictions in Hundreds of Papers Evaluating Graph Representations}
\author {
    Li-Chung Lin,\textsuperscript{\rm 1}
	Cheng-Hung Liu,\textsuperscript{\rm 1}
    Chih-Ming Chen,\textsuperscript{\rm 2}
    Kai-Chin Hsu,\textsuperscript{\rm 3}\\
	I-Feng Wu,\textsuperscript{\rm 4}
	Ming-Feng Tsai,\textsuperscript{\rm 2}
	Chih-Jen Lin\textsuperscript{\rm 1}
}
\affiliations {
    \textsuperscript{\rm 1}National Taiwan University\\
    \textsuperscript{\rm 2}National Chengchi University\\
    \textsuperscript{\rm 3}University of Southern California\\
    \textsuperscript{\rm 4}ASUS Intelligent Cloud Services\\
    r08922141@ntu.edu.tw, ericliu8168@gmail.com, 104761501@nccu.edu.tw, kaichinh@usc.edu,\\
	ifengwu1518@gmail.com, mftsai@nccu.edu.tw, cjlin@csie.ntu.edu.tw
}

\def\bx{{\boldsymbol x}}

\def\by{{\boldsymbol y}}

\def\ovr{{\sf one-vs-rest-basic}\xspace}
\def\ovrc{{\sf one-vs-rest-basic-C}\xspace}
\def\ovrne{{\sf one-vs-rest-no-empty}\xspace}
\def\ovrol{{\sf one-vs-rest-one-label}\xspace}
\def\costsens{{\sf cost-sensitive}\xspace}
\def\costsensne{{\sf cost-sensitive-no-empty}\xspace}
\def\costsenssimp{{\sf cost-sensitive-simple}\xspace}
\def\thresholding{{\sf thresholding}\xspace}
\def\unreal{{\sf unrealistic}\xspace}

\newtheorem{theorem}{Theorem}
\AtEndDocument{\refstepcounter{equation}\label{finaleq}}
\AtEndDocument{\refstepcounter{table}\label{finaltab}}

\begin{document}
\maketitle
	\newcommand*\samethanks[1][\value{footnote}]{\footnotemark[#1]}
	
	\begin{abstract}
		\par Prediction using the ground truth sounds like an oxymoron in machine learning.
		However, such an unrealistic setting was used
		in hundreds, if not thousands of papers in the area of finding graph representations.
		To evaluate the multi-label problem of node classification by using the obtained
		representations, many works assume that the number of
		labels of each test instance is known in the prediction stage.
		In practice such ground truth information is rarely available, but we point out that
		such an inappropriate setting is now ubiquitous in this research area.
		We detailedly investigate why the situation occurs.
		Our analysis indicates that with unrealistic information, the performance is
		likely over-estimated.
		To see why suitable predictions were not used,
		we identify difficulties in applying some multi-label techniques.
		For the use in future studies,
		we propose simple and effective settings 
		without using practically unknown information. 
		Finally, we take this chance to compare
		major graph-representation learning methods
		on multi-label node classification.		
	\end{abstract}
	
	\section{Introduction}
	\label{sec:intro}

	\par Recently unsupervised representation learning over graphs has been an important
	research area.
    One of the primary goals is to find embedding vectors as feature representations of
	graph nodes.
    Many effective techniques \cite[e.g.,][]{BP14a,JT15a,AG16a} have been developed and
	widely applied.
    This research area is very active as can be seen from the tens of thousands of related papers.
    \par The obtained embedding vectors can be used in many downstream tasks, an
	important one being node classification.
    Because each node may be associated with multiple labels,
    this application falls into the category of multi-label problems in machine learning.
    In this study, we point out that in many (if not most) papers using node classification
	to evaluate the quality of
	embedding vectors, an unrealistic setting was adopted for prediction and evaluation.
    Specifically, in the prediction stage, the number of labels of each test instance is
	assumed to be known.
    Then according to decision values, this number of top-ranked labels is considered to
	be associated with the instance.
    Because information on the number of labels is usually 
	not available in practice,
	this setting violates the machine learning principle 
	that ground-truth information should not be used  in the prediction stage.
    Unfortunately, after surveying numerous papers, 
	we find that this inappropriate setting is so ubiquitous that many started thinking it
	is a standard and valid one.
    \par While the research community should move to use appropriate settings,
	some detailed investigation is needed first.
    In this work, we aim to do so by answering the following research questions.
    \begin{itemize}
        \item Knowing this unrealistic setting has been commonly used, how serious is the
		situation and why does it occur?
        \par
		To confirm the seriousness of the situation,
		we identify a long list of papers that have used the
		unrealistic predictions.
        Our analysis then indicates that with unrealistic information,
		the performance is likely over-estimated.
        Further, while the setting clearly cheats, it roughly works for some
		node classification problems that are close to a multi-class one with many
		single-labeled instances.

		\item What are suitable settings without using unknown information?
		Are there practical difficulties 
		for researchers to apply them?

		\par After explaining that multi-label algorithms and/or tools may
		not be readily available, we suggest pragmatic
		solutions for future studies.
		Experimental comparisons with the unrealistic setting
		show that we can effectively optimize some commonly
		used metrics such as Macro-F1.
		\item Because of the use of unrealistic predictions, past comparisons on methods to
		generate embedding vectors
		may need to be re-examined. Can we give
		comparisons under appropriate multi-label predictions?
        \par By using suitable prediction settings,
        our results give new insights into comparing
		influential methods on representation learning.
    \end{itemize}
    \par This paper is organized as follows. Sections
	\ref{sec:issu-exist-sett}-\ref{sec:analysis-unrealistic}
	address the first research question, while Sections
	\ref{sec:appropriate-prediction} and \ref{sec:exp}
	address the second and the third research
	questions, respectively. Finally, Section \ref{sec:conclusion}
	concludes this work. Programs and
	supplementary materials are available at
	\url{www.csie.ntu.edu.tw/~cjlin/papers/multilabel-embedding/}

	\section{Unrealistic Predictions in Past Works}
	\label{sec:issu-exist-sett}
	\par After finding the embedding vectors, past studies on representation learning experiment
	with various applications. An important downstream task is node classification,
	which is often a multi-label classification problem.
	\par In machine learning, multi-label classification is a well-developed area with
	many available training methods. The most used one may be the simple
	one-versus-rest setting, also known as binary relevance.
	This method has been adopted by most works on representation learning.
    The main idea is to train 
	a binary classification problem for each label on data with/without that label.
	The binary optimization problem on label-feature pairs
	$(y_i, \boldsymbol{x}_i),$ where $y_i = \pm 1$ and
	$i=1,\ ...,$ \# training instances, takes the following form.
		\begin{equation}\label{eq:problem}
		\displaystyle\min_{\pmb{w}}\quad \frac{1}{2} \pmb{w}^T \pmb{w} +
		C \sum\nolimits_i \xi(y_i \pmb{w}^T \pmb{x}_i),
		\end{equation}
		where $\xi(\cdot)$ is the loss function, $\pmb{w}^T \pmb{w}/2$
		is the regularization, and $C$ is the regularization
		parameter.\footnote{In some situations a bias term is
		considered, so $\pmb{w}^T \pmb{x}_i$ is replaced by
		$\pmb{w}^T \pmb{x}_i + b$.} Now embedding vectors
		$\pmb{x}_i,\ \forall i$ are available and fixed throughout
		all binary problems. Then for each label, the construction
		of problem \eqref{eq:problem} is simply by assigning
		\begin{equation*}
		y_i=
		\begin{cases}
		1, & \text{if}\ \boldsymbol{x}_i\ \text{is associated with the label}, \\
		-1, & \text{otherwise.}
		\end{cases}
		\end{equation*}
        Because representation learning aims to get a low-dimensional but
		informative vector, a linear classifier is
		often sufficient in the downstream task.
		For the loss function, logistic regression is usually considered, and
		many use the software LIBLINEAR \citep{REF08a} to solve \eqref{eq:problem}.

		\par To check the performance
		after the training process, 
		we find
		that hundreds,
		if not thousands of papers\footnote{See a long list compiled in
		supplementary materials.} in this area used the following
		procedure.
    \begin{itemize}
		\item Prediction stage: for each test instance, assume
	 	\begin{center}
			the number of labels of this instance is known.
		\end{center}
		Predict this number of labels by selecting those with 
		the largest decision values from all binary models.
		\item Evaluation stage: many works report Micro-F1 and Macro-F1.
	\end{itemize}
	Clearly, this setting violates the principle that in the prediction
	stage, ground-truth information should not be used.
	The reason is obvious that in the practical model deployment,
	such information is rarely available.
	\par In particular, some influential works with thousands of citations
	\citep[e.g.,][]{BP14a,JT15a} employed such unrealistic predictions,
	and many subsequent works followed. The practice is now
	ubiquitous and here we quote the descriptions in some papers.
	\begin{itemize}
		\item \citet{SC20a}: ``As in \citet{BP14a} and \citet{JQ18a}, we assume that the number of labels for each test example is given.''
		\item \citet{JS19a}: ``we first obtain the number of actual labels to predict for each sample from the test set. ... This is a common choice in the evaluation setup of the reproduced methods.''
	\end{itemize}
	Interestingly, we find that such unrealistic predictions were used long before the many recent studies on representation learning. An example is as follows.
	\begin{itemize}
    	\item \citet{LT09b}: ``we assume the number of labels of
		unobserved nodes is already known and check the match of the
		top-ranking labels with the truth.''\footnote{\citet{LT09b}
		stated that ``Such a scheme has been adopted for other
		multi-label evaluation works \citep{YL06a}''.
		However, we found no evidence that 
		\citet{YL06a} assumed that the number of labels is known.}
	\end{itemize}
    Our discussion shows how
    an inappropriate setting can eventually propagate to an entire research area.
    Some works did express concerns about the setting.
    For example,
	\begin{itemize}
	    \item \citet{EF18a}: ``Precisely, this method uses the actual number of labels $k$ each test instance has. ... In real world applications, it is fairly uncommon that users have such knowledge in advance.''\footnote{See the version at \url{https://arxiv.org/abs/1710.06520}}
	    \item \citet{XL18a}: ``we note that at the prediction stage previous approaches often employs information that is typically unknown. Precisely, they use the actual number of labels $m$ each testing node has \citep{BP14a,JQ18a}. ... However, in real-world situations it is fairly uncommon to have such prior knowledge of $m$.''
	\end{itemize}
	To be realistic, \citet{EF18a,XL18a} predict labels by checking the sign of 
	decision values.\footnote{More precisely, 
	if logistic regression is used, 
	they check if the probability is greater than 0.5 or not. 
	This is the same as checking the decision value in \eqref{eq:runbinary}.}
	We name this method and give its details as follows.
	\begin{itemize}
		\item \ovr: for a test instance $\bx$,
		\begin{equation}\label{eq:runbinary}
		\pmb{w}^T\pmb{x}
		\begin{dcases}
		\geq 0\\
		< 0
		\end{dcases}
		\Rightarrow
		\begin{dcases}
		\pmb{x}\  \text{predicted to have the label,}\\
		\text{otherwise.}
		\end{dcases}
		\end{equation}
	\end{itemize}
	Their resulting Macro-F1 and Micro-F1 are much lower than works that have used
	unknown information.
	\par If so many works consider an unrealistic setting for predictions, they probably
	have reasons for doing so. Some papers explain the difficulties that lead to their
	assumption of knowing the number of labels.
	\begin{itemize}
		\item \citet{JL16a}: ``As the datasets are not only multi-class but also multi-label,
		we usually need a thresholding method to test the results. But literature gives
		a negative opinion of arbitrarily choosing thresholding methods because of the
		considerably different performances. To avoid this, we assume that the number of
		the labels is already known in all the test processes.'' 
		\item \citet{JQ18a}: ``To avoid the thresholding effect \citep{LT09c}, we assume
		that the number of labels for test data is given \citep{BP14a,LT09c}.''
	\end{itemize}
	To see what is meant by the thresholding effect and the difficulties
	it imposes, we give a simple illustration. For a data
	set BlogCatalog (details in Section \ref{exp:settings}),
	we apply the one-vs-rest training on embedding vectors
	generated by the method DeepWalk \citep{BP14a}. Then the unrealistic
	prediction of knowing the number of labels in each test instance
	is performed. Results (Micro-F1 = 0.41, Macro-F1 = 0.27)
	are similar to those reported in some past works.

	\par In contrast, when using the \ovr setting as in \citet{EF18a,XL18a},
	results are very poor (Micro-F1 = 0.33 and Macro-F1 = 0.19).
	We see that many instances are predicted to have no label at all. 
	A probable cause of this situation is the class imbalance of
	each binary classification problem. That is, in problem
	\eqref{eq:problem}, few training instances have $y_i = 1$,
	and so the decision function tends to predict
	everything as negative. Many multi-label techniques are
	available to address such difficulties, and an important
	one is the thresholding method \citep[e.g.,][]{YY01a,REF07a}.
    Via a constant $\Delta$ to adjust the decision value, in \eqref{eq:runbinary}
	we can replace
	\begin{equation}
		\label{eq:threshold}
	    \pmb{w}^T\pmb{x}\quad \text{with}\quad  \pmb{w}^T\pmb{x} + \Delta.
	\end{equation}
	A positive $\Delta$ can make 
	the binary problem produce more positive predictions.
	Usually $\Delta$ is decided by a cross-validation (CV) procedure.
	Because each label needs one $\Delta$, the overall procedure is more complicated than \ovr.
	Moreover, the training time is significantly longer.
	Therefore, past works may not consider such a technique.

	\section{Analysis of the Unrealistic Predictions}
	\label{sec:analysis-unrealistic}
    We analyze the effect of using the unrealistic predictions.
	To facilitate the discussion,
	in this section we consider
	\begin{equation*}
		i:\text{index of test instances, and }
		j:\text{index of labels.}
	\end{equation*}
	We further assume that for test instance $i$,
	\begin{equation}
		\begin{split}
			&K_i:\text{true number of labels},\\
			&\hat{K}_i:\text{predicted number of labels}.
		\end{split}
	\label{eq:unreal_k_def}
	\end{equation}
	\par In multi-label classification, two types of evaluation metrics
	are commonly used \citep{XZW17a}.
	\begin{itemize}
		\item Ranking measures: examples include precision@K, nDCG@K,
			ranking loss, etc. 
			For each test instance, all we need to predict is
			a ranked list of labels.
		\item Classification measures:
		examples include Hamming loss, Micro-F1, Macro-F1, Instance-F1,
		etc. For each test instance,
		several labels are chosen as the predictions.
	\end{itemize}
	Among these metrics,
	Macro-F1 and Micro-F1 are used in most works on 
	representation learning.
	We first define
	Macro-F1, which is the average of F1 over labels:
	\begin{equation}\label{eq:macro_def}
		\text{Macro-F1}
		=
		\text{Label-F1}
		=
		\frac{\sum \text{F1 of label }j}{\#\text{labels}},
	\end{equation}
	where
	\begin{equation*}
		\text{F1 of label }j = \displaystyle \frac{2\times \text{TP}_j}{\text{TP}_j+\text{FP}_j+\text{TP}_j+\text{FN}_j}.
	\end{equation*}
	Note that $\text{TP}_j$, $\text{FP}_j$, and $\text{FN}_j$
	are respectively the number of true positives,
	false positives and false negatives
	on the prediction of a given label $j$.
	Then Micro-F1 is the F1 by considering
	all instances (or all labels) together:
	\begin{equation}\label{eq:micro}
		\text{Micro-F1}
		=
		\frac{2 \times \text{TP sum}}
		{\text{TP sum + FP sum + TP sum + FN sum}},
	\end{equation}
	where ``sum'' indicates the 
	accumulation of prediction results over all binary problems.
	Next we prove an upper bound of Micro-F1.
	\begin{theorem}\label{th:micro_bound}
		With the definition in \eqref{eq:unreal_k_def},
		we have
		\begin{equation}\label{eq:micro-th}
		\text{Micro-F1}
		\le
		\frac{2 \times \sum\nolimits_{i=1}^{l} 
		\min\bigl(\hat{K}_i,K_i\bigl) }
		{\sum\nolimits_{i=1}^{l} \bigl(K_i + \hat{K}_i\bigl)}
		\le
		1,
		\end{equation}
		where $l$ is the number of test instances.
		Moreover, when $\hat{K}_i = K_i$,
		the bound in \eqref{eq:micro-th} achieves the maximum (i.e., 1).
	\end{theorem}
	The proof is in supplementary materials.
	For the upper bound of Micro-F1 
	proved in Theorem \ref{th:micro_bound},
	we see that knowing
	$K_i$ ``pushes'' the bound to its maximum.
	If a larger upper bound leads to a larger Micro-F1,
	then Theorem \ref{th:micro_bound} indicates the
	advantage of knowing $K_i$.
	
	While Theorem \ref{th:micro_bound}
	proves only an upper bound,
	by some assumption on the decision values,\footnote{
	\citet{XZW17a}
	also assumed \eqref{eq:>eq} for analyzing Micro-F1.
	However, their results are not suited for our
	use here because of various reasons.
	In particular, they made a strong assumption that
	Micro-F1 is equal to Instance-F1.}
	we can exactly obtain Micro-F1 for analysis.
	The following theorem
	shows that
	if all binary models are 
	good enough,
	the upper bound in \eqref{eq:micro-th}
	is attained.
	Further,
	if $K_i$ is 
	known,
	we achieve the best possible $\text{Micro-F1}=1$.
	\begin{theorem}\label{thm:unrealif1}
		Assume for each test instance $i$,
		decision values are properly ranked so that
		\begin{equation}
			\begin{split}
				&\  \text{decision values of its $K_i$ labels}\\
				>&\ 
			\text{decision values of other labels.}
			\end{split}
			\label{eq:>eq}
		\end{equation}
		Under specified $\hat{K}_i$, $\forall$ i,
		the best Micro-F1 is obtained by predicting labels
		with the largest decision values.
		Moreover, the resulting Micro-F1 is the same
		as the upper bound in \eqref{eq:micro-th}.
		That is,
		\begin{equation}\label{eq:perfect-ranking-micro}
		\text{Micro-F1}
		=
		\frac{2 \times \sum\nolimits_{i=1}^{l} 
		\min\bigl(\hat{K}_i,K_i\bigl) }
		{\sum\nolimits_{i=1}^{l} \bigl(K_i + \hat{K}_i\bigl)}.
		\end{equation}
		If $\hat{K}_i = K_i$,
		the best possible $\text{Micro-F1}=1$
		is attained.
	\end{theorem}
	The proof is in supplementary materials.
	Theorem \ref{thm:unrealif1} indicates that even if
	the classifier can output properly ranked decision values,
	without the true number of
	labels $K_i$,
	optimal Micro-F1 still may not be obtained.
	Therefore, using $K_i$
	gives predictions an inappropriate advantage and
	may cause the performance to be over-estimated as a result.

	Next, we investigate why unrealistic predictions were commonly considered
	and point out several possible reasons in the current
	and subsequent sections.
	The first one is 
	the relation to multi-class problems.
	Some popular node classification
	benchmarks are close to multi-class problems
	because many of their instances
	are single-labeled with $K_i = 1$.
	See the data statistics in Table \ref{tab:data}.
	For multi-class problems,
	the number of labels (i.e., one)
	for each instance is known.
	Thus in prediction, we simply find the most probable label.
	In this situation,
	Theorem \ref{th:multiclass} shows that the accuracy commonly used
	for evaluating multi-class problems
	is the same as Micro-F1.
	The proof is in supplementary materials.
	\begin{theorem}
		\label{th:multiclass}
		For multi-class problems,
		\begin{center}
			accuracy = Micro-F1.
		\end{center}
	\end{theorem}
	Therefore, using Micro-F1 with
	prior knowledge on the number of labels
	is entirely valid for multi-class classification.
	Some past studies may conveniently but 
	erroneously extend the setting to multi-label problems.
	\par Based on the findings so far,
	in Section \ref{ssec:predicting-1+} we explain that the unrealistic
	prediction roughly works
	if a multi-label problem contains mostly single-labeled instances.

	\subsection{Predicting at Least One Label per Instance}
	\label{ssec:predicting-1+}
	The discussion in Theorem \ref{th:multiclass} leads to an interesting issue on whether in multi-label classification, at least one label should be predicted for each instance.
	In contrast to multi-class classification,
	for multi-label scenarios,
	we may predict that an instance is associated with no label.
	For the sample experiment on \ovr in Section \ref{sec:issu-exist-sett},
	we mentioned that this ``no label'' situation occurs on many test instances
	and results in poor performance.
	A possible remedy by tweaking
  	the simple \ovr \ method is:
	\begin{itemize}
      \item \ovrne: The method is the same as \ovr,
          except that for instances predicted to have no label,
          we predict the label with the highest decision value.
	\end{itemize}
	For the example considered in Section
	\ref{sec:issu-exist-sett}, this new setting 
	greatly improves the result to 
	0.39 Micro-F1 and 0.24 Macro-F1.
	If we agree that 
	each instance is associated with at least a label (i.e., $K_i \geq 1$), 
	then the method 
	\ovrne \ does not take any unknown information in the prediction stage.
	In this regard, the method of unrealistic predictions is
	probably usable for single-labeled instances.
	However, it is definitely inappropriate 
	for multi-labeled instances.
	For some benchmark sets in Section \ref{sec:exp},
	the majority of instances are multi-labeled.
	Thus there
	is a need to develop effective prediction methods
	without using unrealistic information.
	This subject will be discussed in Section \ref{sec:appropriate-prediction}.

	\section{Appropriate Methods for Training and Prediction}
	\label{sec:appropriate-prediction}
	Multi-label classification is a well-developed area,
	so naturally we may criticize researchers in representation
	learning for not applying suitable techniques.
	However, this criticism may not be entirely fair:
	what if algorithms and/or tools on the multi-label
	side are not quite ready for them?
	In this section, we discuss the difficulties faced by
	researchers on representation learning and explain why
	simple and effective settings are hard to obtain.

	\par The first challenge faced by those handling multi-label
	problems is that they must choose from a myriad of methods
	according to the properties of their applications.
	Typically two considerations are
	\begin{itemize}
	\item number of labels, and
	\item evaluation metrics.
	\end{itemize}
	For example, some problems have extremely many labels,
	and the corresponding research area is called
	``eXtreme Multi-label Learning (XML);'' see the website \citep{KB16a}
	containing many such sets.
	For this type of problems it is impossible to train
	and store the many binary models used by the one-vs-rest setting,
	so advanced methods that organize labels into a tree structure are needed
	\citep[e.g.,][]{RY19a,SK20a,WCC21a}.
	With a huge number of tail labels (i.e., labels that rarely occur), the resulting
	Macro-F1, which is the average F1 over all labels, is often too low to be used.
	In practice, a short ranked list is considered in the
	prediction stage, so precision@K or nDCG@K commonly
	serve as the evaluation metrics.
	\par Nevertheless, the focus now is on node classification problems in past
	studies on representation learning. 
	The number of labels is
	relatively small, and some even contain many
	single-labeled instances.
	From the predominant use of Micro-F1 and Macro-F1 in past works
	it seems that a subset of labels instead of a ranked list is needed
	for node classification.
	Therefore, our considerations are narrowed to 
	\begin{itemize}
	\item methods that are designed for problems without too many labels, and
	\item methods that can predict a subset of labels (instead of just ranks)
	and achieve a high classification measure such as Micro-F1, Macro-F1,
	and Instance-F1.
	\end{itemize}
	\par In addition to one-vs-rest, other methods are applicable
	for our scenario \citep[e.g.,][]{FT12a,JR11a,JR08a,GT07b}.
	Because one-vs-rest does not consider label correlation,
	this aspect is the focus of some methods.
	For simplicity we
	stick with the one-vs-rest setting here
	and prioritize achieving good Macro-F1.
	Macro-F1 in \eqref{eq:macro_def}
	is the average of F1 results over labels,
	so under the one-vs-rest framework,
	all we need is to design a method that can
	give satisfactory F1 on each single label.
	In contrast, optimizing Micro-F1 is more
	difficult because it couples 
	all labels and all instances together;
	see the definition in \eqref{eq:micro}.\footnote{
	See, for example, ``... is the most challenging measure,
	since it does not decompose	over instances nor over labels.''
	in \citet{IP17a}}
	Therefore, we mainly focus on techniques to optimize Macro-F1
	in the following sections.
	
	\subsection{Extending One-vs-rest to Incorporate Parameter Selection}\label{sec:1vr+c_tuning}
	If we examine the \ovr method more closely, it is easy
	to see that a crucial process is missing -- parameter
	selection of the regularization parameter $C$. While the
	importance of parameter selection is well recognized,
	this step is easily forgotten in many places \citep{JJL21a}.
	For example, out of the works that criticized the unrealistic setting
	(see Section \ref{sec:issu-exist-sett}), \citet{EF18a}
	used a fixed regularization parameter for comparing with past works, but \citet{XL18a}
	conducted cross-validation in their one-vs-rest
	implementation. 
	Therefore, a more appropriate baseline
	should be the following extension of \ovr:
	\begin{itemize}
		\item \ovrc: For each binary problem, cross-validation
			is performed on the training data by checking a
			grid of $C$ values. The one yielding the best
			F1 score is chosen to train the binary model
			of the label for future prediction.
	\end{itemize}
	CV is so standard in machine learning that the above
	procedure seems to be extremely simple. Surprisingly,
	several issues may hamper its wide use.
	\begin{itemize}
		\item We learned in Section \ref{sec:issu-exist-sett}
			that some binary problems may not predict any
			positives in the prediction process. Thus
			cross-validation F1 may be zero under
			all $C$ values. In this situation, which
			$C$ should we choose?
		\item To improve robustness, should the same
			splits of data for CV be used throughout
			all $C$ values?
		\item If $C$ is slightly changed from one value
			to another, solutions of the two binary optimization
			problems may be similar. Thus a warm-start implementation
			of using the solution of one problem as the initialization
			for training the other can effectively
			reduce the running time. However, the implementation,
			together with CV, can be complicated.
	\end{itemize}
	The discussion above shows that even for a setting as
	simple as \ovrc, off-the-shelf implementations may
	not be directly available to users.\footnote{
	LIBLINEAR supports warm-start and same CV
	folds for parameter selection after their work in \citet{BYC15a}.
	However, the purpose is to optimize CV accuracy. Our
	understanding is that an extension to check F1 scores
	is available only very recently.}

	\subsection{Thresholding Techniques}\label{sec:thresholding}
	While the basic concept of thresholding has been discussed in Section 2,
	the actual procedure is more complicated and several
	variants exist \citep{YY01a}.
	From early works such as \citet{DL96a,YY99a},
	a natural idea is 
	to use decision values of validation data to decide $\Delta$
	in \eqref{eq:threshold}.
	For each label, the procedure is as follows. 
	\begin{itemize}
		\item For each CV fold, sort validation decision values.
		\par Sequentially assign $\Delta$ as the midpoint
		of two adjacent decision values and
		select the one achieving the best F1 as the
		threshold of the current fold.
		\item Solve a binary problem \eqref{eq:problem} using all training data.
		The average of $\Delta$ values over all folds is then used to adjust the decision function.
	\end{itemize}
	However, \citet{YY01a} showed that this setting easily overfits data 
	if the binary problem is unbalanced.
	Consequently, the same author proposed the $fbr$ heuristic to
	reduce the overfitting problem.
	Specifically, if the F1 of a label is smaller than a pre-defined $fbr$ value,
	then the threshold is set to the largest decision value 
	of the validation data.
	This method requires a complicated two-level
	CV procedure. The outer level uses CV to check that
	among a list of given $fbr$ candidates, 
	which one leads to the best F1.
	The inner CV checks if the validation F1 is better than the given $fbr$.
	\par The above $fbr$ heuristic was further studied
	in an influential paper \citep{DL04b}.
	An implementation from \citet{REF07a} as a LIBLINEAR
	extension has long been publicly available.
	Interestingly, our survey seems to indicate that
	no one in the field of representation learning ever tried it.
	One reason may be that the procedure is complicated.
	If we also select the parameter $C$, then a cumbersome
	outer-level CV to sweep some $(C, fbr)$ pairs is needed.
	Furthermore, it is difficult to use the same data split,
	especially in the inner CV. Another reason may be that
	as a heuristic, people are not confident about the
	method. For example, \citet{LT09b} stated that because
	``thresholding can affect the final prediction performance drastically \citep{REF07a,LT09c},''
	they decided that
	``For evaluation purpose, we assume the number of labels of unobserved nodes is already known.''
	
	\subsection{Cost-sensitive Learning}
	\label{sec:cost-sensitive-learning}
	We learned in Section \ref{sec:issu-exist-sett} that 
	because of class imbalance,
	\ovr suffers from the issue of predicting very few positives.
	While one remedy is the thresholding technique 
	to adjust the decision function,
	another possibility is to conduct cost-sensitive learning.
	Namely, by using a higher loss on positive training instances
	(usually through a larger regularization parameter),
	the resulting model may predict more positives.
	For example, \citet{PP14a} give some theoretical support
	showing that the F1 score can be optimized
	through cost-sensitive learning.
	They extend the optimization problem \eqref{eq:problem} to
	\begin{equation*}
		\displaystyle
		\min_{\pmb{w}} \ \frac{1}{2} \pmb{w}^T \pmb{w} +
		C^+ \!\! \sum\limits_{i:y_i=1} \xi(y_i \pmb{w}^T \pmb{x}_i) +
		C^- \!\!\!\! \sum\limits_{i:y_i=-1} \xi(y_i \pmb{w}^T \pmb{x}_i),
	\end{equation*}
	where
	\begin{equation*}
		C^+ = C(2-t),\ C^- = Ct,\ \text{and } t \in [0, 1].
	\end{equation*}
	Then we can check cross-validation F1 on a grid of
	$(C, t)$ pairs. The best pair is then applied to the whole
	training set to get the final decision function.

	\par An advantage over the thresholding method
	($fbr$ heuristic) is that only a one-level CV is needed.
	However, if many $(C, t)$ pairs are checked, the running
	time can be long. In Section \ref{sec:train-predict-settings}
	we discuss two implementations for this approach.

	\section{Experiments}
	\label{sec:exp}
	In this section we experiment with training/prediction methods
	discussed in Sections
	\ref{sec:issu-exist-sett}-\ref{sec:appropriate-prediction} 
	on popular node classification
	benchmarks. Embedding vectors are generated by some
	well-known methods and their quality is assessed.

	\subsection{Experimental Settings}\label{exp:settings}
	\begin{table}[tb]
          \centering
           \setlength{\tabcolsep}{1pt}          
		\begin{tabular}{@{}c@{}|@{}cc@{\hskip0.2pt}|@{\hskip0.2pt}c@{\hskip0.2pt}|@{\hskip0.2pt}c@{}} %
			\multirow{2}{*}{Data}
			& \multicolumn{2}{c|}{\#instances} & \multirow{2}{*}{\#labels} &
			\multirow{2}{*}{\shortstack[l]{avg. \#labels \\ per instance}}\\
			
			& {\centering single-labeled} & {multi-labeled} & &   \\
			\hline
			BlogCatalog & 7,460 & 2,852 & 39 & 1.40 \\
			Flickr & 62,521 & 17,992 & 195 & 1.34 \\
			YouTube & 22,374 & 9,329 & 46 & 1.60 \\
			PPI & 85 & 54,873 & 121 & 38.26 \\
		\end{tabular}
		\caption{Data statistics.}
		\label{tab:data}
	\end{table}

	We consider the following popular node classification problems:
	\begin{center}
		BlogCatalog, Flickr, YouTube, PPI.
	\end{center}
	From data statistics in Table \ref{tab:data},
	some have many single-labeled instances,
	but some have very few.
	We generate embedding vectors 
	by the
	following influential works.
	\begin{itemize}
		\item DeepWalk \citep{BP14a}.
		\item Node2vec \citep{AG16a}.
		\item LINE \citep{JT15a}.
	\end{itemize}
	Since we consider representation learning independent of
	the downstream task, 
	the embedding-vector generation is unsupervised.
	As such, deciding the parameters for each method can be
	tricky. We reviewed many past
	works and selected the most used values.

	In past studies, Node2vec often had two of its parameters $p,q$
	selected based on the results of the downstream task. 
	This procedure is in effect a
	form of supervised learning.
	Therefore, 
	in our experiments, 
	the parameters $p,q$ are fixed to the
	same values for all data sets.

	For training each binary problem, logistic regression
	is solved by the software LIBLINEAR
	\citep{REF08a}.
	We follow many existing works to randomly
	split each set to $80\%$ for training and $20\%$
	for testing.
	This process is repeated five times and the average score is presented.
	The same training/testing split is used across the
	different graph representations. More details on experimental settings are given
	in the supplementary materials.

	\subsection{Multi-label Training and Prediction Methods for Comparisons}
	\label{sec:train-predict-settings}
	We consider the following methods.
	Unless specified,
	for binary problems \eqref{eq:problem}, we mimic
	many past works to set $C=1$.
	\begin{itemize}
		\item \unreal: After the one-vs-rest training, the unrealistic
			prediction of knowing the number of
			labels is applied.
		\item \ovr: After the one-vs-rest training, 
			each binary classifier
			predicts labels that have positive decision values.
		\item \ovrc: The method,
			described in Section \ref{sec:1vr+c_tuning},
			selects the parameter $C$ by cross-validation.
			We use a LIBLINEAR parameter-selection functionality that
			checks dozens of automatically selected $C$ values.
			It applies a warm-start technique 
			to save the running time.
			An issue mentioned in Section \ref{sec:1vr+c_tuning}
			is that CV F1=0 for every $C$ may occur.
			We checked a few ways to choose $C$
			in this situation, 
			but find results do not differ much.
		\item \ovrne: This method slightly extends \ovr
			so that if all decision values of a test
			instance are negative, then we predict
			the label with the highest decision value;
			see Section \ref{ssec:predicting-1+}.

		\item \thresholding: The method was described in Section \ref{sec:thresholding}. 		
	\end{itemize}
	For the approach in Section \ref{sec:cost-sensitive-learning}
	we consider two variants.
	\begin{itemize}
		\item \costsens: A dense grid of $(C, t)$ is used. The
			range of $t$ is $\{0.1,0.2,\ldots,1\}$. For each
			$t$, we follow \ovrc to use
                        a LIBLINEAR functionality that
			checks dozens of automatically selected $C$ values.
                        In this variant, we do not ensure that
			CV folds are the same across different $t$.
		\item \costsenssimp: We check fewer
			parameter settings by considering $t = \{1/7,2/7,\ldots,1\}$
			and $C = \{0.01/t, 0.1/t, 1/t,10/t,100/t\}$.
			We ensure the
			same data split is applied on the CV for every pair.
			The implementation is relatively simple if
			all parameter pairs are independently trained without
			time-saving techniques such as warm-start.
	\end{itemize}

	Similar to \ovr,
	for \thresholding or \costsens approaches,
	an instance may be predicted to have no labels.
	Therefore, we check the following extension.
	\begin{itemize}
		\item \costsensne: This method extends \costsens by
			the same way from \ovr to \ovrne.
	\end{itemize}

	\begin{table*}[tb]
		\setlength{\tabcolsep}{5pt}
		\centering
		\begin{tabular}{l|ccc|ccc|ccc}
			Training and
			& \multicolumn{3}{c|}{BlogCatalog}
			& \multicolumn{3}{c|}{Flickr}
			& \multicolumn{3}{c}{PPI} \\
			prediction methods
			& DeepWalk & Node2vec & LINE
			& DeepWalk & Node2vec & LINE
			& DeepWalk & Node2vec & LINE \\
			\hline
			\multicolumn{3}{c}{} & \multicolumn{7}{c}{Macro-F1 \;\;(avg. of five; std. in supplementary)} \\
			\hline
			\unreal & 0.276 & 0.294 & 0.239 & 0.304 & 0.306 & 0.258 & 0.483 & 0.442 & 0.504 \\
			\cline{2-10}
			\ovrc & 0.208 & 0.220 & 0.195 & 0.209 & 0.208 & 0.188 & 0.183 & 0.150 & 0.243 \\
			\thresholding & 0.269 & 0.283 & 0.221 & {\bf0.299} & {\bf0.302} & 0.264 & {\bf0.482} & 0.457 & {\bf0.498} \\
			\costsens & {\bf0.270} & {\bf0.283} & {\bf0.250} & 0.297 & 0.301 & {\bf0.279} & 0.482 & {\bf0.461} & 0.495 \\
			\hline
			\multicolumn{3}{c}{} & \multicolumn{7}{c}{Micro-F1 \;\;(avg. of five; std. in supplementary)} \\
			\hline
			\unreal & 0.417 & 0.426 & 0.406 & 0.416 & 0.420 & 0.409 & 0.641 & 0.626 & 0.647 \\
			\cline{2-10}
			\ovrc & 0.344 & 0.355 & 0.335 & 0.291 & 0.296 & 0.289 & 0.458 & 0.441 & 0.489 \\
			\thresholding & {\bf0.390} & {\bf0.396} & {\bf0.353} & {\bf0.370} & {\bf0.376} & {\bf0.364} & {\bf0.535} & 0.482 & {\bf0.553} \\
			\costsens & 0.366 & 0.371 & 0.341 & 0.352 & 0.358 & 0.354 & 0.533 & {\bf0.495} & 0.548 \\
		\end{tabular}
		\caption{Results of representative training/prediction
		methods applied to various embedding vectors.
		Each value is the average of five 80/20 training/testing
		splits.
		The score of the best training/prediction method (excluding \unreal)
		is bold-faced.}
		\label{tab:expbase}
	\end{table*}

	\begin{table*}[tb]
		\setlength{\tabcolsep}{3.93pt}
		\centering
		\begin{tabular}{l|cc|cc|cc|cc}
			Training and prediction
			& \multicolumn{2}{c|}{BlogCatalog}
			& \multicolumn{2}{c|}{Flickr}
			& \multicolumn{2}{c|}{YouTube}
			& \multicolumn{2}{c}{PPI} \\
			methods on DeepWalk vectors
			& Macro-F1 & Micro-F1
			& Macro-F1 & Micro-F1
			& Macro-F1 & Micro-F1
			& Macro-F1 & Micro-F1 \\
			\hline
			\ovr & 0.190 & 0.334 & 0.195 & 0.283 & 0.213 & 0.287 & 0.181 & 0.449 \\
			\ovrc & 0.208 & 0.344 & 0.209 & 0.291 & 0.217 & 0.290 & 0.183 & 0.458 \\
			\ovrne & 0.241 & {\bf 0.390} & 0.256 & {\bf 0.377} & 0.263 & {\bf 0.382} & 0.181 & 0.449 \\
			\hline
			\costsens & {\bf 0.270} & 0.366 & 0.297 & 0.352 & {\bf 0.360} & 0.374 & {\bf 0.482} & {\bf 0.533} \\
			\costsensne & 0.268 & 0.351 & {\bf 0.298} & 0.343 & 0.359 & 0.372 & {\bf 0.482} & {\bf 0.533} \\
			\costsenssimp & 0.266 & 0.353 & 0.297 & 0.358 & 0.357 & 0.372 & 0.481 & 0.529 \\
		\end{tabular}
		\caption{Ablation study on variations of \ovr and \costsens
		applied to embedding vectors generated by DeepWalk.
		Each value is the average of five 80/20 training/testing
		splits. The best training/prediction method is bold-faced.}
		\label{tab:expext}
	\end{table*}

	\subsection{Results and Analysis}
	In Table \ref{tab:expbase} we compare the \unreal method and
	representative methods in Section \ref{sec:appropriate-prediction}.
	Other variants are investigated in Table \ref{tab:expext} later.
	Due to the space limit,
	we omit the YouTube data set, though results follow similar trends.
	Observations from Table \ref{tab:expbase} are as follows.
	\begin{itemize}
		\item As expected, \unreal is the best in nearly all situations.
		It significantly outperforms others on Micro-F1, 
		a situation confirming not only 
		the analysis in Theorem \ref{th:multiclass}
		but also that \unreal may over-estimate performance.
		\item In Section \ref{sec:issu-exist-sett} we showed an example that
		\ovr performs poorly because of the thresholding issue.
		Even with the parameter selection, \ovrc still suffers
		from the same issue and performs the worst.
		\item Both \thresholding and \costsens effectively optimize
		Macro-F1 and achieve similar results to \unreal. Despite Micro-F1
		not being the optimized metric, the improvement over
		\ovrc is still significant.
	\end{itemize}

	In Table \ref{tab:expext} we study the variations of \ovr and \costsens.
	We only present the results of embedding vectors generated by DeepWalk,
	while complete results with similar trends
	are in supplementary materials.
	Some observations 
	from Table \ref{tab:expext}
	are as follows.
	\begin{itemize}
		\item Even with parameter selection, \ovrc is only
		marginally better than \ovr. This result is possible
		because for binary logistic regression,
		it is proved that after $C$ is sufficiently large,
		the decision function is about the same
		(Theorem 3 in \citealp{BYC15a}). The result shows
		that conducting parameter selection is not enough
		to overcome the thresholding issue.
		\item Following the analysis in Section \ref{ssec:predicting-1+},
		\ovrne significantly improves upon \ovr
		for problems that have many single-labeled instances.
		However, it has no visible effect on the set PPI,
		in which most instances are multi-labeled.
		\item However, \costsensne shows no such improvement over \costsens
		because \costsens mitigates the issue of predicting no
		labels for a large portion of instances. Further, for the
		remaining instances with no predicted labels, the
		label with the highest decision value may be an incorrect one,
		resulting in worse Micro-F1 in some cases.
		This experiment shows
		the importance to have techniques that allow empty predictions.
		\item \costsenssimp is generally competitive with
		\costsens and \thresholding.
	\end{itemize}

	\par An issue raised in Section \ref{sec:appropriate-prediction}
	is whether the same split of data (i.e., CV folds)
	should be used in the multiple CV procedures ran by, for example,
	\costsenssimp. We have conducted some analysis, but leave
	details in supplementary materials due to the space limitation.

	Regarding methods for representation learning,
	we have the following observations.
	\begin{itemize}
		\item Our results of the \unreal method are close
		to those in the recent comparative study \citep{MK21a}.
		This outcome supports the validity of our experiments.
		\item Among the three methods to generate representations,
		there is no clear winner,
		indicating that the selection may be application dependent.
		DeepWalk and Node2vec are closer to each other
		because they are both based on random walks.
		In contrast, LINE is based on edge modeling.
		\item DeepWalk is a special case of Node2vec
		under some parameter values,
		though here Node2vec is generated by other commonly suggested values.
		Because DeepWalk is generally competitive
		and does not require the selection of some Node2vec's parameters,
		DeepWalk may be a better practical choice.
		\item The relative difference between the three
		representation learning methods differs from what \unreal
		suggests. Even though in our comparisons such effects
		are not large enough to change their relative ranking,
		an unfair comparison diminishes the utility of
		benchmark results.
	\end{itemize}
	
	\section{Conclusions}\label{sec:conclusion}
	We summarize the results on training/prediction methods.
	The two methods \thresholding
	and \costsens are effective and can be applied in future studies.
	They are robust without the concerns mentioned in some papers. Further,
	if an easy implementation is favored, then the simple yet competitive
	\costsenssimp can be a pragmatic choice.
	The implementations are available in an easy-to-use package
	\centerline{\url{https://github.com/ASUS-AICS/LibMultiLabel}}
	Thus, researchers in the area of representation learning
	can easily apply appropriate prediction settings.

	\par In the well-developed world of machine learning,
	it may be hard to believe that unrealistic predictions were
	used in almost an entire research area.
	However, it is not the time to blame
	anyone. Instead, the challenge is to ensure that
	appropriate settings are used in the future. In this
	work, we analyze how and why unrealistic predictions were
	used in the past. We then discuss suitable replacements.
	Through our investigation hopefully
	unrealistic predictions will no longer be used.

	\section{Acknowledgments}
	This work was supported by MOST of Taiwan grant 110-2221-E-002-115-MY3
	and ASUS Intelligent Cloud Services.

	\bibliography{aaai22}

\end{document}